\documentclass[10pt,twocolumn,letterpaper]{article}

\usepackage{cvpr}              
\usepackage[accsupp]{axessibility} 

\usepackage[dvipsnames]{xcolor}

\definecolor{cvprblue}{rgb}{0.21,0.49,0.74}
\usepackage[pagebackref,breaklinks,colorlinks,citecolor=cvprblue]{hyperref}

\def\paperID{7899} 
\def\confName{CVPR}
\def\confYear{2024}

\title{Puff-Net: Efficient Style Transfer with Pure Content and Style Feature Fusion Network}

\author{Sizhe Zheng, ~Pan Gao\thanks{Corresponding authors.}, ~Peng Zhou, ~Jie Qin\footnotemark[1] \\
College of Computer Science and Technology, Nanjing University of Aeronautics and Astronautics\\
{\tt\small {\{162100101, pan.gao, zhoupeng23, jie.qin\}}@nuaa.edu.cn} }

\begin{document}
\maketitle
\begin{abstract}
Style transfer aims to render an image with the artistic features of a style image, while maintaining the original structure. Various methods have been put forward for this task, but some challenges still exist. For instance, it is difficult for CNN-based methods to handle global information and long-range dependencies between input images, for which transformer-based methods have been proposed. Although transformers can better model the relationship between content and style images, they require high-cost hardware and time-consuming inference. To address these issues, we design a novel transformer model that includes only the encoder, thus significantly reducing the computational cost. In addition, we also find that existing style transfer methods may lead to images under-stylied or missing content. In order to achieve better stylization, we design a content feature extractor and a style feature extractor, based on which pure content and style images can be fed to the transformer. Finally, we propose a novel network termed Puff-Net, i.e., \textbf{pu}re content and style \textbf{f}eature \textbf{f}usion \textbf{net}work. Through qualitative and quantitative experiments, we demonstrate the advantages of our model compared to state-of-the-art ones in the literature.
The code is available at \href{https://github.com/ZszYmy9/Puff-Net} {https://github.com/ZszYmy9/Puff-Net}.
\end{abstract}    
\section{Introduction}
\label{sec:intro}

As personalized expression gains popularity, people increasingly seek to transform images into new artistic styles. Imagine turning a plain landscape photo into an oil painting or a snapshot into an Impressionist-inspired image. This technology, known as Style Transfer in computer vision, offers possibilities for artistic expression. It captures and blends the essence of artistic styles into different images, creating pieces that merge original content with artistic styles.

\begin{figure}[!t]
    \centering
    \includegraphics[width=\linewidth]{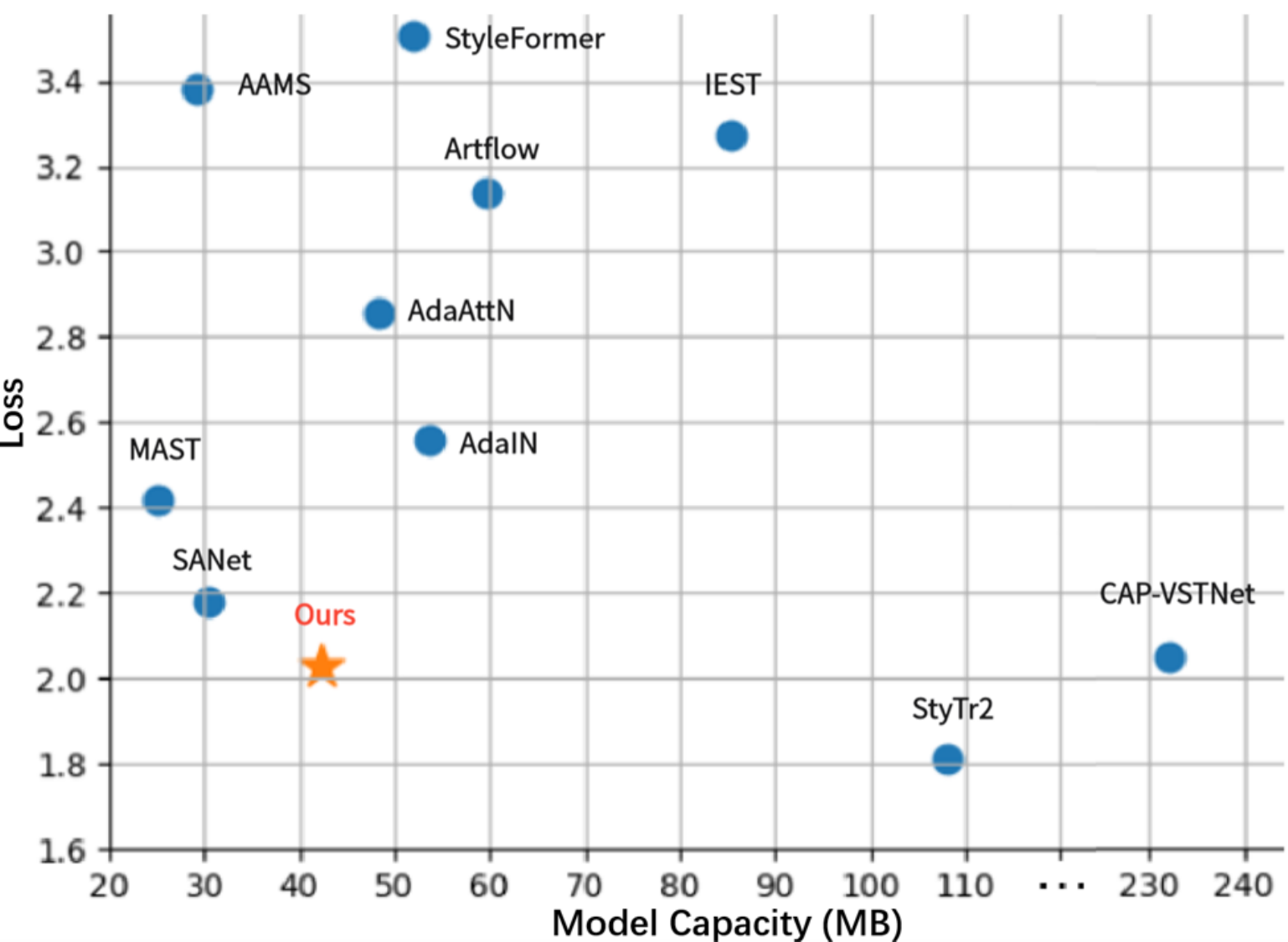}
    \caption{Comparison of different models based on loss and capacity, with the loss being a combination of $60\%$ content loss and $40\%$ style loss. Our model shows a favorable balance between capacity and loss. Details can be found in the Method and Experiments sections.}
    \label{Figure:comparison}
    \vspace{-5mm}
\end{figure}

Early style transfer methods primarily relied on optimization algorithms which minimize the differences between the input image and the reference image. However, the high computational complexity of these methods greatly limited their practical applications. With technological advancements, image style transfer techniques based on direct inference have made significant progress. The method introduced by Gatys \emph{et al.}~\cite{gatys2016image}, which employs convolutional neural networks (CNN), extracts features of content and style from different layers of a pre-trained CNN model. This approach has significantly reduced computational complexity and spurred a wave of related research, including developments like AdaIN~\cite{huang2017arbitrary}, Avatar~\cite{sheng2018avatar}, SANet~\cite{park2019arbitrary}, and MAST~\cite{deng2020arbitrary}. Despite the achievements of these CNN-based inference methods for image style transfer, they still face limitations. They depend on convolution operations to capture image features, and their performance is limited when the network layers are insufficient to capture global information. On the other hand, as the number of layers increases, the content details of the synthesized image may be lost, which in turn affects the overall quality of the stylized image. Therefore, effectively transferring style while maintaining content integrity remains a challenge in the field of image style transfer.

\begin{figure}[!t]
    \centering
    \includegraphics[width=\linewidth]{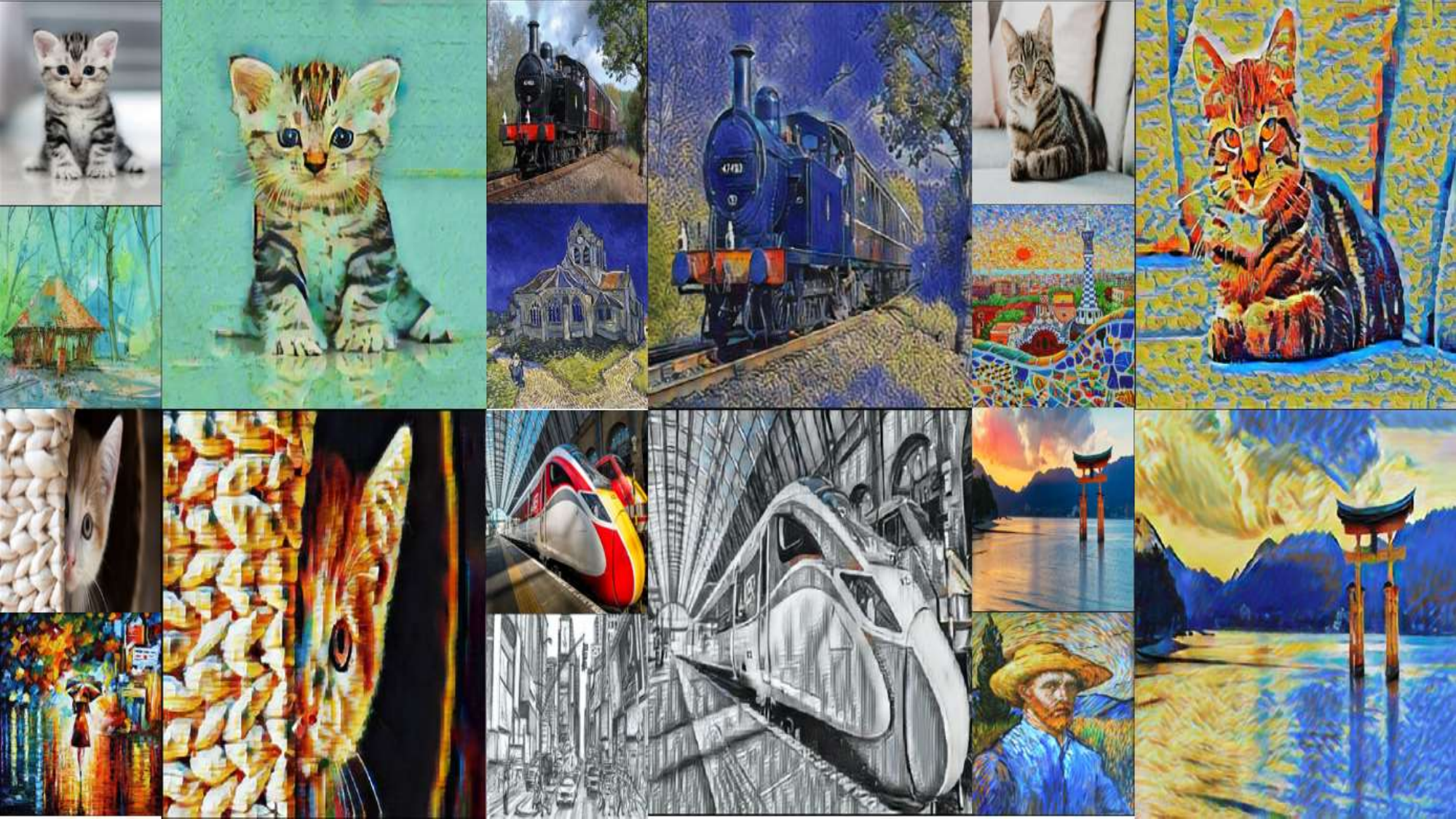}
    \caption{Some results of our Puff-Net. Our method achieves a better balance between maintaining stylized effects and reducing computational costs. The main body and background of the content image can be stylized more reasonably based on the style image. 
    }
    \label{Figure:result}
    \vspace{-3mm}
\end{figure}

\begin{figure*}[!t]
    \centering
    \includegraphics[width=\linewidth]{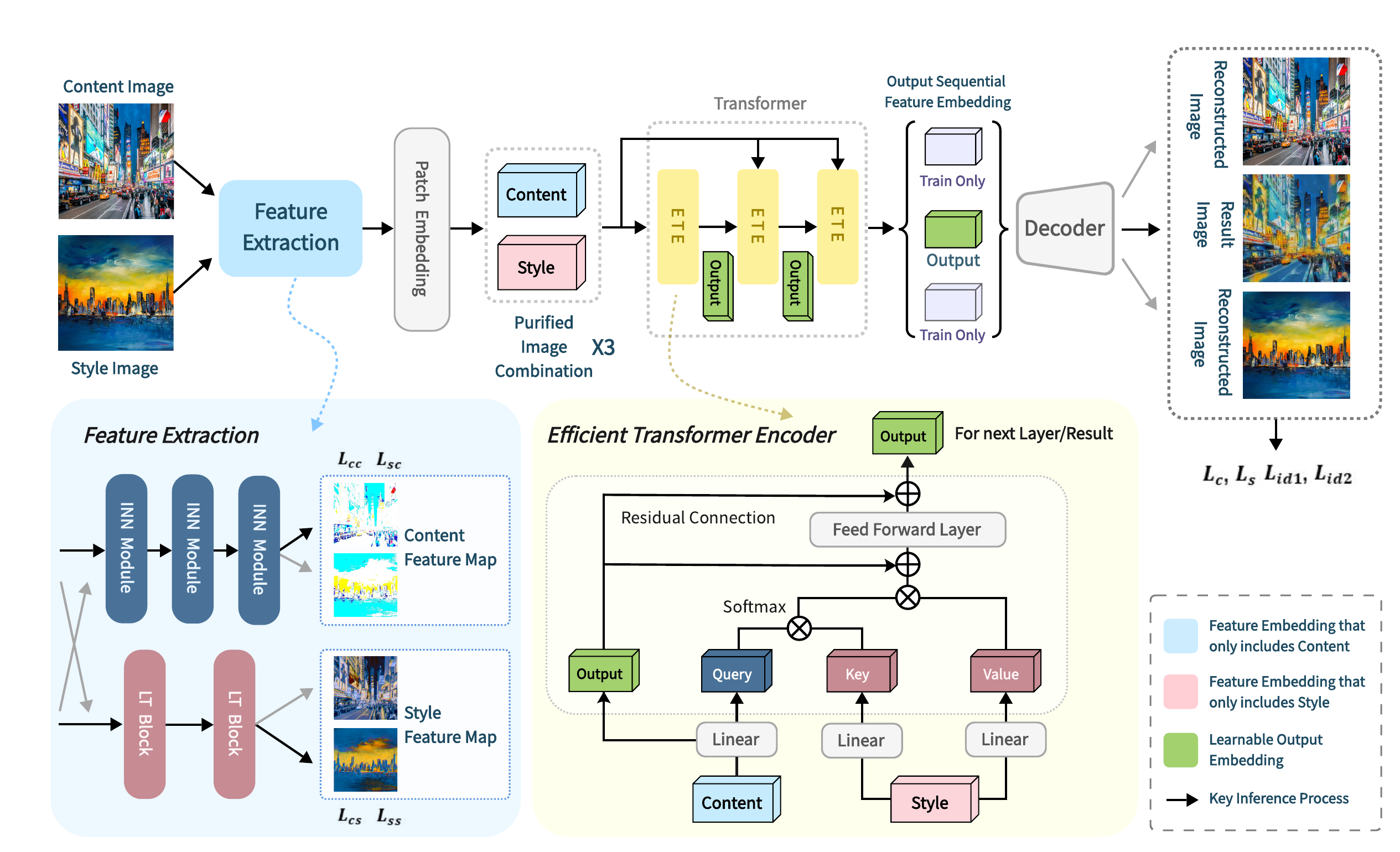}
    \vspace{-6mm}
    \caption{Schematic illustration of the Puff-Net architecture. The network begins by extracting content and style features from the input images. These features are then divided into patches and encoded into patch sequences through a linear projection. After feeding the features into the transformer for stylization, we can finally obtain the result image through the decoder. Additionally, the model leverages a reconstruction loss function during training to enhance its ability to reconstruct content and style features. }
    \label{Fig:Workflow}
    \vspace{-5mm}
\end{figure*}

Vision transformer (ViT)~\cite{dosovitskiy2020image} offers a novel approach for utilizing transformer models~\cite{vaswani2017attention} in visual tasks. By dividing input images into a series of small patches and rearranging them to form embedding vectors, this method has been shown to surpass the performance of traditional CNNs in several visual tasks. For example, a transformer-based model~\cite{deng2022stytr2} has achieved significant breakthroughs in style transfer tasks. The success of transformers in handling image data can be attributed to their attention mechanism, which captures the global context of images. Additionally, this mechanism helps the model better understand the relationship between content and style in images, enhancing stylization efforts.
However, the model capacity of transformer is large, with high hardware requirements and slow training speed. In order to tackle these difficulties, we design a transformer that includes only the encoder. We modify the encoder structure of the transformer so that we can obtain stylized output sequences of image patches through the encoder alone. The modified transformer is of lower complexity and has a significantly improved inference speed. 

By analyzing the generated results, we found that the generated images may have significant differences from the content images. The feature distribution of the generated images is not visually reasonable where the content features of some style images also appear. Our objective is to eliminate the style attributes from the content images, preserving only their content structure. Simultaneously, we hope that style images can focus less on content details and allow their style features to participate in the stylization process. Therefore, we preprocess the content images and style images before feeding them to the transformer for stylization. Accordingly, we develop two distinct feature extractors: one to isolate content features and the other to isolate style features from the input images. 

In summary, we introduce a novel framework for efficient style transfer, namely \textbf{pu}re content and style \textbf{f}eature \textbf{f}usion \textbf{net}work (Puff-Net), which incorporates two feature extractors and a transformer equipped solely with an encoder. Our major contributions are summarized as follows: 
\begin{itemize}
    \item We enhance the structure of the encoder in the vanilla transformer so that style transfer can be performed efficiently through only the encoder, reducing computational overhead.
    \item We design two feature extractors that preprocess the input to obtain pure content images and pure style images, and consequently, achieving superior stylized results.
    \item Even with a notable reduction in model capacity, our model continues to deliver competitive performance over existing counterparts.
\end{itemize}

\Cref{Figure:comparison} illustrates the comparison of our model with state-of-the-art models in terms of model capacity and overall loss, and \Cref{Figure:result} provides visual stylized results of our Puff-Net. 
It is evident that the proposed Puff-Net achieves a balance between style transfer effectiveness and model efficiency.
\section{Related Work}
\label{sec:related work}
\subsection{Style Transfer}
Image style transfer has made significant progress. Gatys \emph{et al.}~\cite{gatys2016image} discover that when feeding an input image into a pre-trained CNN (VGG19), one can capture the content and style information of the image and integrate both information by using an optimization-based method. Then, relevant ensuing studies started emerging. AdaIN \cite{huang2017arbitrary} aligns the mean and variance of content image features with the mean and variance of style images to implement style transfer. SANet \cite{park2019arbitrary} integrates local style patterns efficiently and flexibly based on the semantic spatial distribution of content images. MAST \cite{deng2020arbitrary} enhances feature representations of content images and style images through position-wise self-attention, calculates their similarity, and rearranges the distribution of these representations. ArtFlow \cite{an2021artflow} proposes a model consisting of reversible neural flows and an unbiased feature transfer module, which can prevent content leak during universal style transfer. AdaAttN \cite{liu2021adaattn} designs a novel attention and normalization module and inserts it into the traditional encoder-decoder pipeline. IEST \cite{chen2021artistic} utilizes external information and employs contrastive learning for style transfer. StyleFormer \cite{wu2021styleformer} incorporates transformer components into the traditional CNN workflow. StyTr$^2$ \cite{deng2022stytr2} proposes a model which achieves the style trasnsfer only through the vanilla transformer. CAP-VSTNet \cite{wen2023cap} adopts a reversible framework to protect content images to avoid artifacts, and achieves style transfer through an unbiased linear transform module. However, it places more emphasis on retaining content, which leads to the image under-stylized. CLIPstyler \cite{kwon2022clipstyler} achieves style transfer through text injection. We aim to provide a practical solution for style transfer. Puff-Net balances the output quality and efficiency, while CNN-based methods prioritize speed. StyTr$^2$ focuses on higher-quality outputs at the cost of efficiency, and CAP-VSTNet retains more original content details compromising quality. 
More recently, diffusion-based models \cite{ho2020denoising} prioritize creativity over efficiency.

\vspace{-2mm}
\subsection{Transformer}
The vanilla transformer \cite{vaswani2017attention} is designed to tackle tasks in the field of natural language processing (NLP). The unique self-attention mechanism can effectively model the relationships between tokens. In order to apply the transformer to the field of computer vision (CV), lots of related research has been carried out. The proposal of the ViT \cite{dosovitskiy2020image} made a groundbreaking contribution to the application of transformers in CV. It segments images into patches and arranges them into embeddings, which are fed to ViT for processing. Since the advent of ViT, variants of transformers have been proposed to deal with multiple visual tasks. For example, DETR \cite{carion2020end} and YOLOS \cite{fang2021you} for object detection, SegFormer \cite{xie2021segformer} and SETR \cite{zheng2021rethinking} for semantic segmentation, and CTrans \cite{lanchantin2021general} and Swin-Transformer \cite{liu2021swin} for image classification. Transformers are also proven very effective in the area of multi-modal fusion, \emph{e.g.}, ViLT \cite{kim2021vilt}. StyTr$^2$ \cite{deng2022stytr2} adopts only the vanilla transformer for style transfer for the first time, and the improvement is very significant. Compared with the CNN, transformers can capture long-range dependencies of input images by using attention mechanisms.
In this paper, we also leverage the strong global modeling capability of transformers for style transfer. However, different from prior models, we utilize the transformer encoder to associate the disentangled content feature and disentangled style feature, resulting in better stylizing results with a smaller model scale. 

\section{Method}\label{sec:method}

In this section, we will introduce the workflow of the proposed Puff-Net. We set the dimensions of the input and output to be $H \times W \times3$. To make use of the transformer encoder, we treat style transfer as a sequential patch generation task. We split both content and style images into patches and use a linear projection layer to project input patches into a sequential feature embedding $\varepsilon$ in the shape of $L\times C$, where $L = \frac{H \times W}{m \times m}$ is the length of $\varepsilon$, $m = 8$ is the patch size, and $C$ is the dimension of $\varepsilon$. Figure \ref{Fig:Workflow} illustrates the overall architecture of the proposed Puff-Net.

\subsection{Efficient Transformer Encoder}

StyTr$^2$~\cite{deng2022stytr2} employs the transformer to implement the task of style transfer. However, their proposed model necessitates a large number of computational resources. In addition to the encoder, a transformer decoder is adopted to translate the encoded content sequence according to the encoded style sequence. The output sequential feature embedding $\varepsilon_o$ can only be obtained through a complete transformer. This is very cumbersome, and thus we hope to obtain $\varepsilon_o$ directly through the encoder alone. With this objective in mind, we modify the transformer encoder as follows. We append a learnable sequence feature embedding $\varepsilon_o$ whose shape is the same as $\varepsilon_c$, and we process it in the encoder based on the purified content and style images. The FFN layer of the transformer consumes much of the computation, but its role in capturing context features is not significant. To this end, we make it process and transmit the information of $\varepsilon_o$. Through the encoder, we can obtain the required sequence feature embedding $\varepsilon_o$. After applying the decoder, we can obtain the result image. The overview of the Efficient Transformer Encoder (ETE) is shown in Figure \ref{Fig:Workflow}.

Considering that the output image should be close to the content image, we initialize the learnable $\varepsilon_o$ based on $\varepsilon_c$. We intend to connect $\varepsilon_c$ and $\varepsilon_s$ and feed them to the encoder. Each layer of the encoder consists of a multi-head self-attention module (MSA) and a feed-forward network (FFN). However, its computational complexity is $O ( (2L)^2 \times C + 2L \times C^2)$. Therefore, we redesign the transformer model. $\varepsilon_c$ is encoded into a query (Q) and $\varepsilon_s$ is encoded into a key (K) and a value (V). The computational complexity is $O (L^2 \times C + L \times C^2)$. We can also better build the connection between the content and style images in this way. Moreover, when we use the attention mechanism, the positional encoding should be included in the input. Here, we use Content-Aware Positional Encoding (CAPE) in \cite{deng2022stytr2}, which takes image semantics into account when implementing positional encoding. We only calculate CAPE for the content image as follows:
\begin{equation}
    \begin{aligned}
    Q = (\varepsilon_c + \mathcal{P} _{CA}) W_q, K = \varepsilon_s W_k, V = \varepsilon_s W_v
    \end{aligned}
\end{equation}
where $W_q, W_k, W_v \in R^{ C\times d_{head}}$. The multi-head attention is then calculated by
\begin{equation}
    \begin{aligned}
    \mathcal{F}_{MSA}& = Concat (head_1, ……, head_n) W_o;\\
    &head_i = Attention_i (Q, K, V)
    \end{aligned}
\end{equation}
where $W_o \in R^{C\times C}$, $n$ is the number of attention heads, and $d_{head} = \frac{C}{N}$. In each encoder layer, the output $Y$ is calculated as
\begin{equation}
    \begin{aligned}
    &Y' = \mathcal{F}_{MSA}(Q, K, V) + \varepsilon_o;\\
    &Y = \mathcal{F}_{FFN}(Y') + Y'
    \end{aligned}
\end{equation}
where $\mathcal{F}_{FFN}(Y') = max (0, Y'W_1+b_1) W_2 + b_2$. Layernorm~\cite{ba2016layer} is applied in each layer.

Through the encoder, we can obtain the output sequence in the form of $\frac{H \times W}{m \times m} \times C$. To obtain the final result, we employ a three-layer CNN decoder proposed in~\cite{zheng2021rethinking}. In each layer, we expand the scale by executing a series of operations including 3 $\times$ 3 Conv + ReLU + 2 $\times$ Upsample. Finally, we can save the resultant image in the form of H $\times$ W $\times$ 3.

\subsection{Feature Extraction}

Since we aim to transfer the style of the content image, we hope that the underlying model can change the color and other characteristics. Meanwhile, we ought to avoid damaging content images as much as possible. Therefore, we try to increase the proportion of the content loss when designing the loss function. Doing so may result in a small difference between the results and content images, yet still including style features such as color. In order to ensure that the content is not missing while the style of the result image is closer to that of the style image, we preprocess the content and style images to extract their distinct features.
To extract different kinds of features from the content and style images, we assume the two images contain information from two modalities, from which we can capture their unique features. Therefore, we handle the two types of images separately to obtain pure content and style images. Towards this end, we employ different feature extractors for content images and style images, respectively.

To process content images with minimal loss of detail, we select the INN module \cite{dinh2016density} as the backbone for our content extractor, aiming for utmost content preservation.
It can better preserve the content by making its input and output features mutually generated. Therefore, we adopt the INN block \cite{dinh2016density,zhou2022pan} with affine coupling layers. In each invertible layer, the transformation is written as follows:
\begin{equation}
    \begin{aligned}
        Y_{k+1}[c+1:C] &= Y_{k}[c+1:C] + \phi_1(Y_{k}[1:c])\\
        Y_{k+1}[1:c] &= Y_{k}[1:c]\odot \exp(\phi_2(Y_{k+1}[c+1:C])) \\&+ \phi_3(Y_{k+1}[c+1:C])\\
        Y_{k+1} &= Concat(Y_{k+1}[1:c], Y_{k+1}[c+1:C])
    \end{aligned}
\end{equation}
where $\odot$ is the Hadamard product, $Y_k$ ($k$=1,2,$\cdots$) is the output of the $k$-th layer, [1:$c$] represents the $1$st to the $c$-th channels, and $\phi_i$ ($i$=1,2,3) are the arbitrary mapping functions. To balance the feature extraction ability and computational complexity, we employ the bottleneck residual block (BRB) in MobileNetV2 \cite{sandler2018mobilenetv2}.

For style images, our focus is on capturing the general style, rather than the local details.
It requires the extractor to grasp the global information and long-distance dependency features well. Meanwhile, considering the computational complexity of the model, we choose the LT block \cite{2020Lite} as the basic unit of the style extractor. It flattens the bottleneck of transformer blocks by flattening the feed-forward network, which saves substantial computation.
Please refer to the supplementary material for the network details of the two extractors.

\vspace{-3mm}
\subsection{Loss Function}
The generated image requires a fusion of content and style. Therefore, we need a content loss function and a style loss function, respectively. Following \cite{huang2017arbitrary, an2021artflow}, we obtain feature maps through a pretrained VGG model and use them to construct the content perceptual loss $\mathcal{L}_c$ and the style perceptual loss $\mathcal{L}_s$ as follows:
\begin{equation}
    \begin{aligned}
        \mathcal{L}_c = &\frac{1}{N_l}\sum_{i=0}^{N_l}\parallel \psi_i(I_o) -\psi_i(I_c) \parallel_2\\
        \mathcal{L}_s = &\frac{1}{N_l}\sum_{i=0}^{N_l}\parallel \mu(\psi_i(I_o)) - \mu(\psi_i(I_s)) \parallel_2\\&+ \parallel \sigma(\psi_i(I_o)) - \sigma(\psi_i(I_s))\parallel_2\\
    \end{aligned}
\end{equation}
where $I_o$ represents the output of the model, $I_c$ is the content image and $I_s$ is the style image, $\psi_i(\cdot)$ denotes the features extracted from the $i$-th layer in a pretrained VGG19, and $N_l$ is the number of layers. $\mu(\cdot)$ and $\sigma(\cdot)$ denote the mean and variance of the extracted features, respectively.

For the feature extraction module, we can train the two extractors with these loss functions. We adopt the content perceptual loss for the input and output of the content extractor ($\mathcal{L}_{cc}$: the content perceptual loss w.r.t. the content image; $\mathcal{L}_{sc}$: the content perceptual loss w.r.t the style image), and the style perceptual loss for the input and output of the style extractor ($\mathcal{L}_{cs}$: the style perceptual loss w.r.t. the content image; $\mathcal{L}_{ss}$: the style perceptual loss w.r.t. the style image). $\mathcal{L}_{fe}$ is used to calculate the total loss of feature extractors. In order to enhance the learning ability of the extractor, we implement two extractors on both the content and style images. We reconstruct the result image through the content and style features extracted from the same image, and the result image should be consistent with the original image. Here we employ two identity losses \cite{park2019arbitrary} to increase the severity of the penalty as follows:
\begin{equation}
    \begin{aligned}
        \mathcal{L}_{fe} = &\lambda_1\mathcal{L}_{cc} + \lambda_2\mathcal{L}_{cs} + \lambda_1\mathcal{L}_{sc} + \lambda_2\mathcal{L}_{ss}\\
        \mathcal{L}_{id1} = &\parallel I_{cc} - I_c \parallel_2 + \parallel I_{ss} - I_s \parallel_2\\
        \mathcal{L}_{id2} = &\frac{1}{N_l}\sum_{i=0}^{N_l}\parallel \psi_i(I_{cc}) - \psi_i(I_c) \parallel_2\\&+ \parallel \psi_i(I_{ss}) - \psi_i(I_s) \parallel_2
    \end{aligned}
\end{equation}
where $\lambda_1,\lambda_2$ are the weights set to 0.7 and 1, respectively. $I_{cc}$ and $I_{ss}$ are the reconstructed images.

In summary, the entire network is optimized by minimizing the following function:
\begin{equation}
    \begin{aligned}
        \mathcal{L} = \lambda_c \mathcal{L}_c + \lambda_s \mathcal{L}_s + \lambda_{fe} \mathcal{L}_{fe} + \lambda_{id1} \mathcal{L}_{id1} + \lambda_{id2} \mathcal{L}_{id2}.
    \end{aligned}
\end{equation}
We set $\lambda_c,\lambda_s,\lambda_{fe},\lambda_{id1},\lambda_{id2}$ to 7, 10, 20, 70, 1 so as to alleviate the impact from magnitude differences.

\begin{figure*}[!t]
    \centering
    \includegraphics[width=\linewidth]{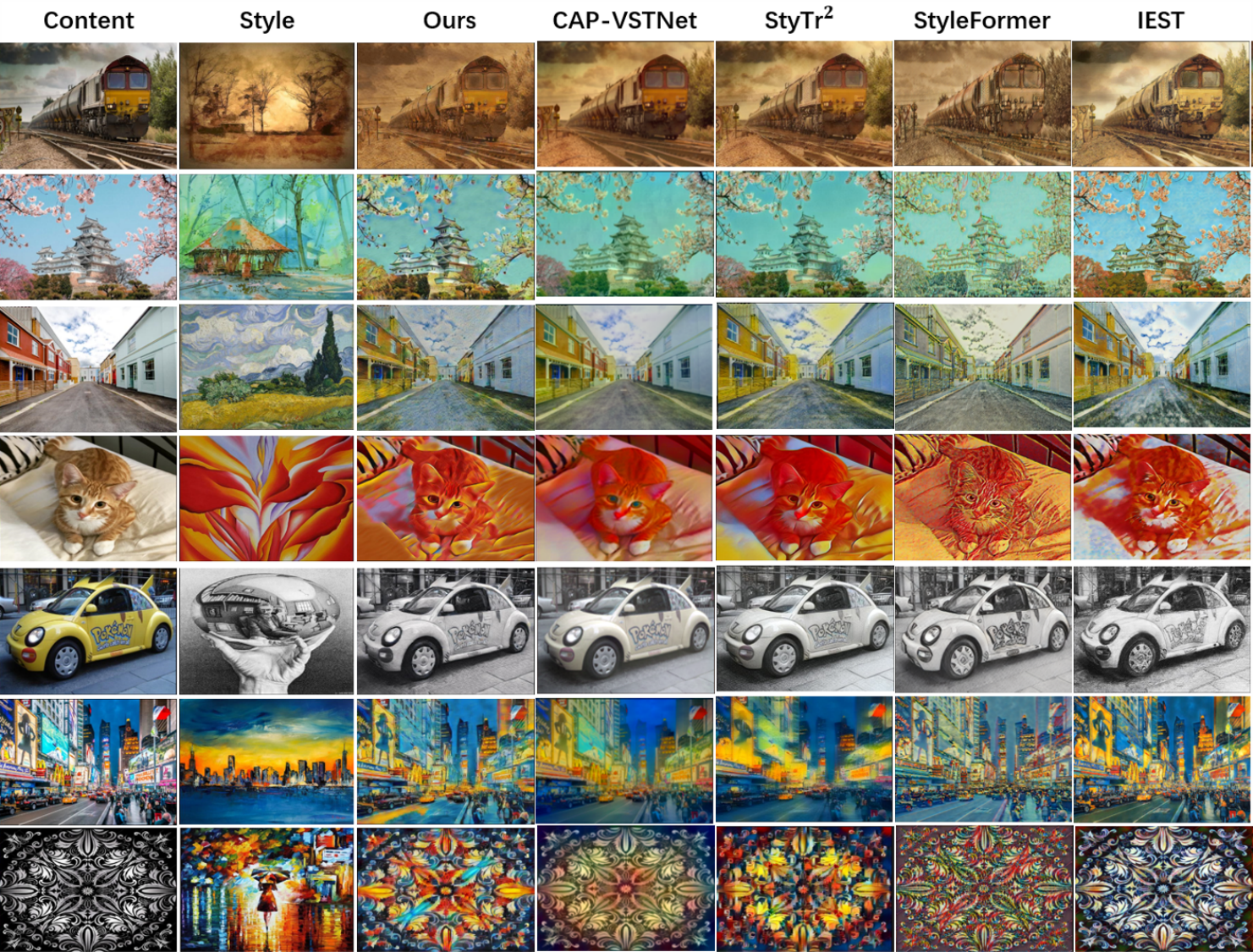}
    \vspace{-7mm}
    \caption{The visual results of qualitative comparisons}
    \label{Figure:Comparison}
    \vspace{-5mm}
\end{figure*}
\section{Experiments}\label{sec:4}

\subsection{Implementation Details}

We adopt MS-COCO \cite{lin2014microsoft} as the content dataset and WikiArt \cite{phillips2011wiki} as the style dataset. In the training stage, all the images are randomly cropped into a fixed resolution of 256 $\times$ 256, while any image resolution is supported at the test time. We choose the Adam optimizer \cite{kingma2014adam} with a learning rate of 0.0005 and use the warm-up adjustment strategy \cite{xiong2020layer}. The batch size is set to 1 and we train our network with 100,000 iterations. Our model is trained on the NVIDIA Tesla A40 for about half a day. 

During the training stage, we found that the style extractor trained for 12,000 iterations produced the best image style. Some style features may disappear after more iterations. We believe this is because the difference between the result image and the content image accounts for a larger proportion of the total loss,  as we tend to preserve content details as much as possible. In order to reduce the total loss, the extracted style features will decrease after more rounds of training. Without stylization, the content perceptual loss will be very low, and so will the total loss. Therefore, we freeze the parameters of the style extractor after 12,000 iterations of training, while the other parts continue to participate in the training. Maybe we can also use two-stage training scheme~\cite{li2021rfn}.

\subsection{Comparison with State-of-the-Art Methods}

Transformer networks have proven their powerful performances in numerous computer vision fields. So far, state-of-the-art models, such as StyTr$^2$ \cite{deng2022stytr2}, have utilized the attention mechanism. CNN-based models, despite their fast inference, can result in missing details due to their limitations of kernel weight sharing. We have chosen the mainstream style transfer models CAP-VSTNet \cite{wen2023cap}, StyTr$^2$ \cite{deng2022stytr2},  StyleFormer \cite{wu2021styleformer}, and IEST \cite{chen2021artistic} for comparison. We conduct both qualitative and quantitative comparisons.

\subsubsection{Qualitative comparison}
Figure \ref{Figure:Comparison} shows the visual results of the qualitative comparisons. The IEST retains more content features, but the degree of stylization is insufficient. CAP-VSTNet also has the same drawback, but its original content is more protected. The StyleFormer sometimes exaggerates details, resulting in some unreasonable stylization. The local details of some results generated by the StyTr$^2$ are not obvious, leading to the content missing. By contrast, our model can effectively utilize the content and style features of input images and exploit their relationships. It can extract the main content lines of the original image and adopt attention mechanism to stylize these main features, which can maintain the global structure of the content image and make the stylized image look very coordinated. But we also observe that when the input content images and style images are more complex, sometimes there may be unreasonable stylization.

\begin{table}[!t]
    \centering
    \setlength{\tabcolsep}{4.5pt}
    \tabcolsep=0.1cm
        \begin{tabular}{c|ccccc}
            \toprule
             Resolution & Ours &CAP-VST& StyTr$^2$ & StyleFormer & IEST \\ 
             \midrule
             256 $\times$ 256 & 0.098 & 0.107 & 0.116 & 0.013 & 0.065 \\ 
             \midrule
             512 $\times$ 512 & 0.134 & 0.162 & 0.661 & 0.026 & 0.092 \\ 
             \bottomrule
    \end{tabular}%
    \vspace{-3mm}
    \caption{Average inference time (in seconds) of the comparison methods at two output resolutions.}
    \label{tab:time}%
\end{table}

\subsubsection{Quantitative comparison}
In Table \ref{tab:time}, we compare the inference time of these models at two output resolutions using one NVIDIA Tesla P100. As can be seen from the table, our model's inference speed is at the forefront of these mainstream models. 

To quantitatively analyze the effect of generating  stylized images, we randomly select 20 content images and 20 style images, and then use the mainstream models to generate 200 stylized images. We calculate the content difference and the style difference using (5). Table \ref{tab:loss} shows that our model's comprehensive performance is at the forefront. In terms of content difference, we have a small gap compared to the StyTr$^2$. Our method also achieves the second-lowest style loss. Although the style difference is slightly greater, we do not pay much attention to the local detail differences between the result image and style image. What's more, we can see that CAP-VSTNet has the lowest content loss, but its degree of stylization is lacking. Through the quantitative analysis, one can see that our proposed model still retains a good performance despite significantly reduced model capacity. This provides a better method for the application of style transfer.

\begin{table}[!t]
    \centering
    \setlength{\tabcolsep}{4.5pt}
    \tabcolsep=0.15cm
        \begin{tabular}{c|ccccc}
            \toprule
             Model &Ours&CAP-VST& StyTr$^2$  &StyleFormer & IEST \\ 
             \midrule
             $\mathcal{L}_c$ &1.92&0.86& 1.89 & 2.87 & 1.97 \\ 
             \midrule
             $\mathcal{L}_s$ &2.21&4.42& 1.69 & 3.34 & 3.99  \\ 
             \bottomrule
    \end{tabular}%
    \vspace{-3mm}
    \caption{Quantitative comparison on the content perceptual loss and style perceptual loss.}
    \label{tab:loss}%
\end{table}

\begin{figure}[!t]
    \centering
    \includegraphics[width=\linewidth]{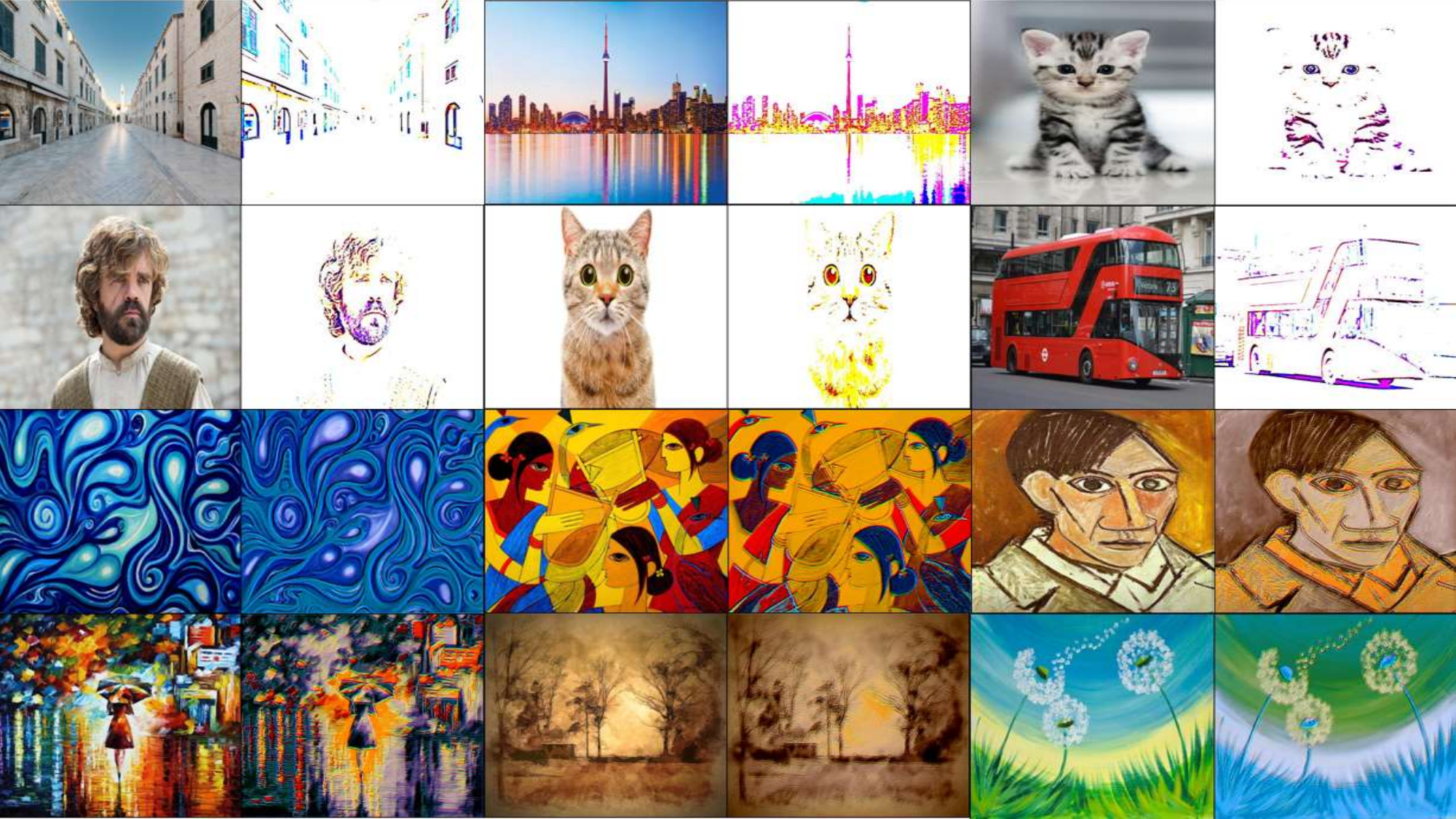}
    \caption{The features extracted by our model. The first and second rows display renderings of the content extractor, while the third and fourth rows display renderings of the style extractor. }
    \label{fig:feature}
    \vspace{-3mm}
\end{figure}

\subsection{Ablation Experiment}

\subsubsection{Feature Extraction}
Attention mechanisms are proven to be effective in the field of style transfer. We intend to investigate whether the feature extractors work. Figure \ref{fig:feature} plots the results generated by these extractors. By using a content feature extractor, the structure, lines, and other content features of an input image are extracted, which meet our expectations. Some background and less important contents are blurred. Through a style feature extractor, the color, texture, and other aspects of the input image are extracted. Although the feature distribution of some images has changed, we do not pay attention to the details of the style image.

\begin{figure}[!t]
    \centering
    \includegraphics[width=\linewidth]{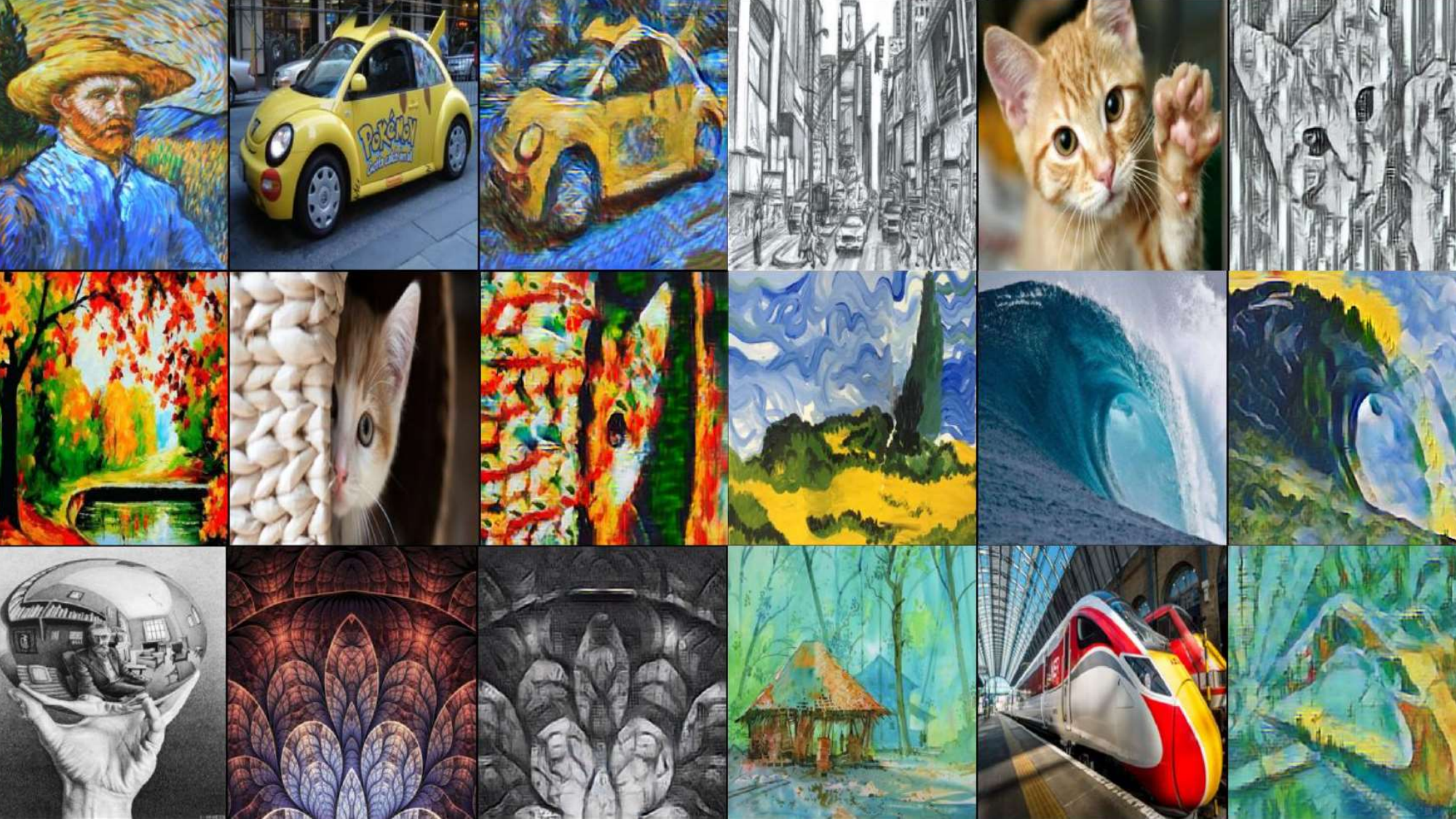}
    \caption{The results without extractors. Each group has three images, the first one is a style image, the second one is a content image, and the last one is the result image. }
    \label{fig:res_with_ab}
    \vspace{-3mm}
\end{figure}

In order to further investigate the efficacy of the feature extractors, we conduct ablation experiments. We remove the content feature extractor and style feature extractor from the model and do not extract the pure features of the content and style images. We directly project their patches into sequential feature embeddings and feed them into the encoder-based transformer. In order to offset the impact of different network depths, we add the encoder layers from 3 to 6. Figure \ref{fig:res_with_ab} shows the results generated by pure encoder without those extractors. 

As can be observed from Figure \ref{fig:res_with_ab}, some generated images have lost their original content structure, resulting in visual distortions (second row, first group). There are also some content features such as lines in the style image appearing in the resultant image (first row, second group). Some images have insufficient stylization in terms of details (third row, second group). Therefore, it can be concluded that our content and style feature extractors are of good performance.

\subsubsection{Content-Aware Positional Encoding}
\begin{figure}[ht]
    \centering
    \includegraphics[width=\linewidth]{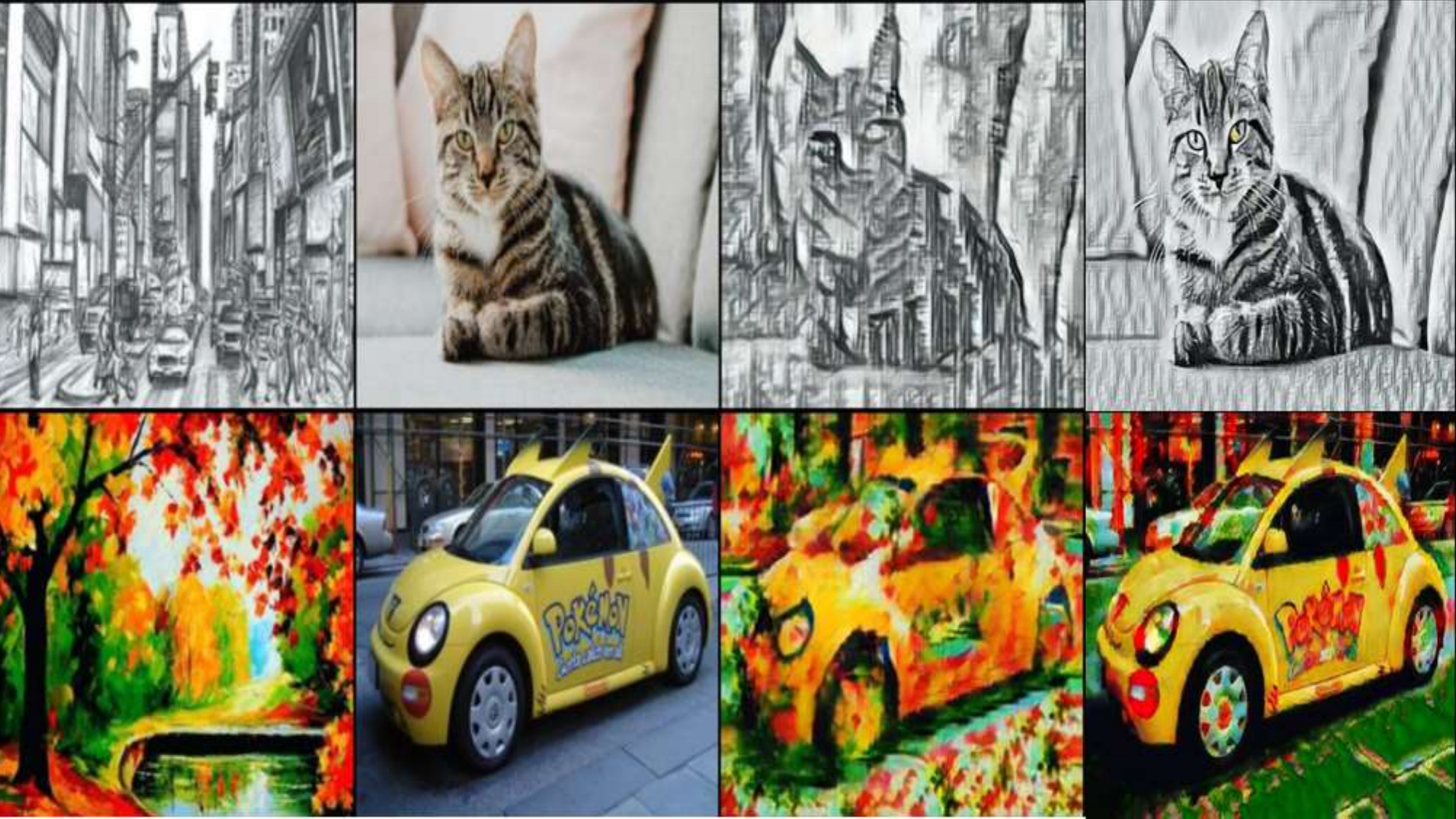}
    \caption{Ablation experiments for CAPE. From the first to the last column: style images, content images, result images using sinusoidal positional encoding, and result images using CAPE.}
    \label{fig:CAPE}
    \vspace{-2mm}
\end{figure}

Content-Aware Positional Encoding (CAPE) is a learnable position encoding method based on image semantic information proposed by \cite{deng2022stytr2}. Since our model extracts features from the content picture image, the content image will lose some semantic information. We have already shown some extracted feature maps in Figure \ref{fig:feature}. Through the content extractor, we can extract the structure, lines, and other features of the input image, but we discard style features such as colors, which destroys some of the semantic information of the image. Therefore, in order to verify whether CAPE can play a better role in our model, we carry out a ablation study. In the ablation experiment, we replace CAPE with traditional sinusoidal positional encoding and trained the model. We present the results of this ablation experiment in Figure \ref{fig:CAPE}.
As can be seen from the experimental results in Figure \ref{fig:CAPE}, the results using CAPE are better than those using traditional sinusoidal positional encoding. The results without using CAPE may have unreasonable stylization, and some originally similar areas may have significant differences after stylization. We believe that although extracting features may cause losses to the semantic information of the image, we still need positional encoding to exploit the remaining information for stylization. Some features such as the background need to be similarly stylized, and different detail features can be stylized differently. Therefore, we still employ CAPE as the positional encoding method.

\subsection{Output Sequential Feature Embedding}

In our model, we can obtain output sequential feature embedding $\varepsilon_o$ only through the encoder. So, we modify the transformer encoder, appending a learnable sequence feature embedding $\varepsilon_o$ to the input. Its shape is the same as content sequential feature embedding $\varepsilon_c$. During the training stage, as we know the resultant image should be closer to the content image, we use $\varepsilon_c$ to initialize it. In order to further investigate its role in the model, we experiment with other initialization methods.

We first initialize it using style sequence feature embedding $\varepsilon_s$. It can be observed that the resultant image is very similar to the style image, which does not meet expectations.
We believe that our model realize stylization based on $\varepsilon_o$, using the attention mechanism to calculate each part's stylization approach. Since we use $\varepsilon_s$ to initialize $\varepsilon_o$, it is in the stylized state from the beginning, and the subsequent stylization effect will not be significant.
We also use random initialization and zero initialization, and find that the generated stylized images are blank. We believe that our model cannot find a suitable way to stylize images without the content.
The qualitative results using different output feature embedding initialization methods are presented in supplementary material due to space constraint.

In summary, $\varepsilon_o$ is the basis for style transfer in our model, and the calculation results of the attention mechanism determine the way of stylization for each patch. Therefore, we choose to initialize it with $\varepsilon_c$, which is more in line with the goal of style transfer.

\subsection{User Study}
\begin{figure}[!t]
    \centering
    \includegraphics[width=\linewidth]{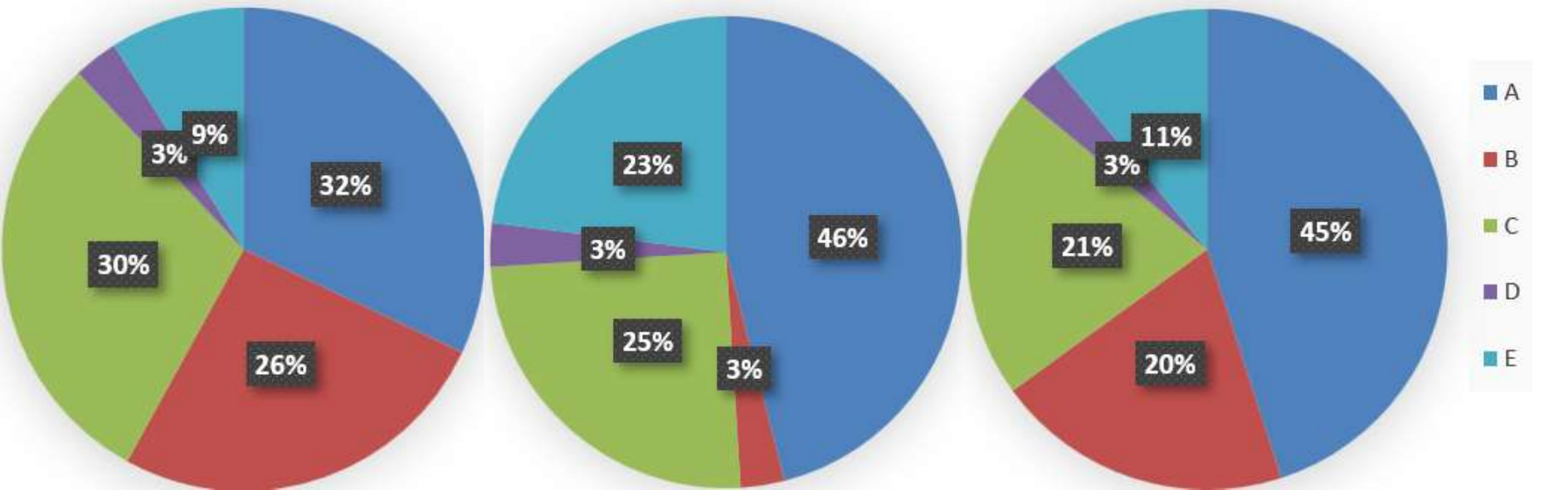}
    \caption{Results of User Study. The above three figures correspond to questions one, two, and three, respectively.
    A-Puff-Net. B-CAP-VSTNet. C-StyTr$^2$. D-StyleFormer. E-IEST. }
    \label{fig:user_study}
    \vspace{-4mm}
\end{figure}

In order to better evaluate the performance of our model, we conduct a user study. The comparison resources come from Figure \ref{Figure:Comparison}. We invited 45 college students and 10 middle-aged people to conduct this survey. We have set three types of questions for the purpose of style transfer task. The first question is which model can better maintain the original image's content structure. The second question is which model's result is closer to the target style image. The third question is which model's result after stylization looks the most harmonious. We will provide two examples for each type of question. The results of the survey are shown in Figure \ref{fig:user_study}. In terms of the example we provided, from the results, we can see that our model's ability to maintain the original image content structure is similar to that of CAP-VSTNet, and its ability to achieve stylization is optimal, followed closely by StyTr$^2$ and IEST. As for the ability to achieve reasonable stylization, our model is also outstanding. In order to further demonstrate the performance of the model and reduce randomness, we hope that more people can use our model to produce the expected results.

\section{Conclusion}\label{sec:5}
In this paper, we proposed a novel style transfer model dubbed the Puff-Net. The proposed model consists of two feature extractors and a transformer that only contains the encoders. We first obtained pure content images and pure style images through the two feature extractors. Then we fed them into an efficient encoder-based transformer for stylization, in which a sequence of learnable tokens were added to interact with pure content and style tokens. Our model solves the problem of huge capacity in existing transformer-based models. We also verified its good performance through extensive experiments and demonstrated the potential application of style transfer in practice.

\section*{Acknowledgement}
This work was partially supported by the National Natural Science Foundation of China (No. 62276129 \& 62272227), and the Natural Science Foundation of Jiangsu Province (No. BK20220890).
{
    \small
    \bibliographystyle{ieeenat_fullname}
    \bibliography{main}

\begin{thebibliography}{32}
\providecommand{\natexlab}[1]{#1}
\providecommand{\url}[1]{\texttt{#1}}
\expandafter\ifx\csname urlstyle\endcsname\relax
  \providecommand{\doi}[1]{doi: #1}\else
  \providecommand{\doi}{doi: \begingroup \urlstyle{rm}\Url}\fi

\bibitem[An et~al.(2021)An, Huang, Song, Dou, Liu, and Luo]{an2021artflow}
Jie An, Siyu Huang, Yibing Song, Dejing Dou, Wei Liu, and Jiebo Luo.
\newblock Artflow: Unbiased image style transfer via reversible neural flows.
\newblock In \emph{Proceedings of the IEEE/CVF Conference on Computer Vision and Pattern Recognition}, pages 862--871, 2021.

\bibitem[Ba et~al.(2016)Ba, Kiros, and Hinton]{ba2016layer}
Jimmy~Lei Ba, Jamie~Ryan Kiros, and Geoffrey~E Hinton.
\newblock Layer normalization.
\newblock \emph{arXiv preprint arXiv:1607.06450}, 2016.

\bibitem[Carion et~al.(2020)Carion, Massa, Synnaeve, Usunier, Kirillov, and Zagoruyko]{carion2020end}
Nicolas Carion, Francisco Massa, Gabriel Synnaeve, Nicolas Usunier, Alexander Kirillov, and Sergey Zagoruyko.
\newblock End-to-end object detection with transformers.
\newblock In \emph{European conference on computer vision}, pages 213--229. Springer, 2020.

\bibitem[Chen et~al.(2021)Chen, Wang, Zhang, Zuo, Li, Xing, Lu, et~al.]{chen2021artistic}
Haibo Chen, Zhizhong Wang, Huiming Zhang, Zhiwen Zuo, Ailin Li, Wei Xing, Dongming Lu, et~al.
\newblock Artistic style transfer with internal-external learning and contrastive learning.
\newblock \emph{Advances in Neural Information Processing Systems}, 34:\penalty0 26561--26573, 2021.

\bibitem[Deng et~al.(2020)Deng, Tang, Dong, Sun, Huang, and Xu]{deng2020arbitrary}
Yingying Deng, Fan Tang, Weiming Dong, Wen Sun, Feiyue Huang, and Changsheng Xu.
\newblock Arbitrary style transfer via multi-adaptation network.
\newblock In \emph{Proceedings of the 28th ACM international conference on multimedia}, pages 2719--2727, 2020.

\bibitem[Dinh et~al.(2016)Dinh, Sohl-Dickstein, and Bengio]{dinh2016density}
Laurent Dinh, Jascha Sohl-Dickstein, and Samy Bengio.
\newblock Density estimation using real nvp.
\newblock \emph{arXiv preprint arXiv:1605.08803}, 2016.

\bibitem[Dosovitskiy et~al.(2020)Dosovitskiy, Beyer, Kolesnikov, Weissenborn, Zhai, Unterthiner, Dehghani, Minderer, Heigold, Gelly, et~al.]{dosovitskiy2020image}
Alexey Dosovitskiy, Lucas Beyer, Alexander Kolesnikov, Dirk Weissenborn, Xiaohua Zhai, Thomas Unterthiner, Mostafa Dehghani, Matthias Minderer, Georg Heigold, Sylvain Gelly, et~al.
\newblock An image is worth 16x16 words: Transformers for image recognition at scale.
\newblock \emph{arXiv preprint arXiv:2010.11929}, 2020.

\bibitem[et~al.(2022)]{deng2022stytr2}
Deng et al.
\newblock Stytr2: Image style transfer with transformers.
\newblock In \emph{CVPR}, pages 11326--11336, 2022.

\bibitem[et~al.(2020)]{ho2020denoising}
Ho et al.
\newblock Denoising diffusion probabilistic models.
\newblock \emph{NIPS}, 33:\penalty0 6840--6851, 2020.

\bibitem[Fang et~al.(2021)Fang, Liao, Wang, Fang, Qi, Wu, Niu, and Liu]{fang2021you}
Yuxin Fang, Bencheng Liao, Xinggang Wang, Jiemin Fang, Jiyang Qi, Rui Wu, Jianwei Niu, and Wenyu Liu.
\newblock You only look at one sequence: Rethinking transformer in vision through object detection.
\newblock \emph{Advances in Neural Information Processing Systems}, 34:\penalty0 26183--26197, 2021.

\bibitem[Gatys et~al.(2016)Gatys, Ecker, and Bethge]{gatys2016image}
Leon~A Gatys, Alexander~S Ecker, and Matthias Bethge.
\newblock Image style transfer using convolutional neural networks.
\newblock In \emph{Proceedings of the IEEE conference on computer vision and pattern recognition}, pages 2414--2423, 2016.

\bibitem[Huang and Belongie(2017)]{huang2017arbitrary}
Xun Huang and Serge Belongie.
\newblock Arbitrary style transfer in real-time with adaptive instance normalization.
\newblock In \emph{Proceedings of the IEEE international conference on computer vision}, pages 1501--1510, 2017.

\bibitem[Kim et~al.(2021)Kim, Son, and Kim]{kim2021vilt}
Wonjae Kim, Bokyung Son, and Ildoo Kim.
\newblock Vilt: Vision-and-language transformer without convolution or region supervision.
\newblock In \emph{International Conference on Machine Learning}, pages 5583--5594. PMLR, 2021.

\bibitem[Kingma and Ba(2014)]{kingma2014adam}
Diederik~P Kingma and Jimmy Ba.
\newblock Adam: A method for stochastic optimization.
\newblock \emph{arXiv preprint arXiv:1412.6980}, 2014.

\bibitem[Kwon and Ye(2022)]{kwon2022clipstyler}
Gihyun Kwon and Jong~Chul Ye.
\newblock Clipstyler: Image style transfer with a single text condition.
\newblock In \emph{Proceedings of the IEEE/CVF Conference on Computer Vision and Pattern Recognition}, pages 18062--18071, 2022.

\bibitem[Lanchantin et~al.(2021)Lanchantin, Wang, Ordonez, and Qi]{lanchantin2021general}
Jack Lanchantin, Tianlu Wang, Vicente Ordonez, and Yanjun Qi.
\newblock General multi-label image classification with transformers.
\newblock In \emph{Proceedings of the IEEE/CVF Conference on Computer Vision and Pattern Recognition}, pages 16478--16488, 2021.

\bibitem[Li et~al.(2021)Li, Wu, and Kittler]{li2021rfn}
Hui Li, Xiao-Jun Wu, and Josef Kittler.
\newblock Rfn-nest: An end-to-end residual fusion network for infrared and visible images.
\newblock \emph{Information Fusion}, 73:\penalty0 72--86, 2021.

\bibitem[Lin et~al.(2014)Lin, Maire, Belongie, Hays, Perona, Ramanan, Doll{\'a}r, and Zitnick]{lin2014microsoft}
Tsung-Yi Lin, Michael Maire, Serge Belongie, James Hays, Pietro Perona, Deva Ramanan, Piotr Doll{\'a}r, and C~Lawrence Zitnick.
\newblock Microsoft coco: Common objects in context.
\newblock In \emph{Computer Vision--ECCV 2014: 13th European Conference, Zurich, Switzerland, September 6-12, 2014, Proceedings, Part V 13}, pages 740--755. Springer, 2014.

\bibitem[Liu et~al.(2021{\natexlab{a}})Liu, Lin, He, Li, Wang, Li, Sun, Li, and Ding]{liu2021adaattn}
Songhua Liu, Tianwei Lin, Dongliang He, Fu Li, Meiling Wang, Xin Li, Zhengxing Sun, Qian Li, and Errui Ding.
\newblock Adaattn: Revisit attention mechanism in arbitrary neural style transfer.
\newblock In \emph{Proceedings of the IEEE/CVF international conference on computer vision}, pages 6649--6658, 2021{\natexlab{a}}.

\bibitem[Liu et~al.(2021{\natexlab{b}})Liu, Lin, Cao, Hu, Wei, Zhang, Lin, and Guo]{liu2021swin}
Ze Liu, Yutong Lin, Yue Cao, Han Hu, Yixuan Wei, Zheng Zhang, Stephen Lin, and Baining Guo.
\newblock Swin transformer: Hierarchical vision transformer using shifted windows.
\newblock In \emph{Proceedings of the IEEE/CVF international conference on computer vision}, pages 10012--10022, 2021{\natexlab{b}}.

\bibitem[Park and Lee(2019)]{park2019arbitrary}
Dae~Young Park and Kwang~Hee Lee.
\newblock Arbitrary style transfer with style-attentional networks.
\newblock In \emph{proceedings of the IEEE/CVF conference on computer vision and pattern recognition}, pages 5880--5888, 2019.

\bibitem[Phillips and Mackintosh(2011)]{phillips2011wiki}
Fred Phillips and Brandy Mackintosh.
\newblock Wiki art gallery, inc.: A case for critical thinking.
\newblock \emph{Issues in Accounting Education}, 26\penalty0 (3):\penalty0 593--608, 2011.

\bibitem[Sandler et~al.(2018)Sandler, Howard, Zhu, Zhmoginov, and Chen]{sandler2018mobilenetv2}
Mark Sandler, Andrew Howard, Menglong Zhu, Andrey Zhmoginov, and Liang-Chieh Chen.
\newblock Mobilenetv2: Inverted residuals and linear bottlenecks.
\newblock In \emph{Proceedings of the IEEE conference on computer vision and pattern recognition}, pages 4510--4520, 2018.

\bibitem[Sheng et~al.(2018)Sheng, Lin, Shao, and Wang]{sheng2018avatar}
Lu Sheng, Ziyi Lin, Jing Shao, and Xiaogang Wang.
\newblock Avatar-net: Multi-scale zero-shot style transfer by feature decoration.
\newblock In \emph{Proceedings of the IEEE conference on computer vision and pattern recognition}, pages 8242--8250, 2018.

\bibitem[Vaswani et~al.(2017)Vaswani, Shazeer, Parmar, Uszkoreit, Jones, Gomez, Kaiser, and Polosukhin]{vaswani2017attention}
Ashish Vaswani, Noam Shazeer, Niki Parmar, Jakob Uszkoreit, Llion Jones, Aidan~N Gomez, {\L}ukasz Kaiser, and Illia Polosukhin.
\newblock Attention is all you need.
\newblock \emph{Advances in neural information processing systems}, 30, 2017.

\bibitem[Wen et~al.(2023)Wen, Gao, and Zou]{wen2023cap}
Linfeng Wen, Chengying Gao, and Changqing Zou.
\newblock Cap-vstnet: Content affinity preserved versatile style transfer.
\newblock In \emph{Proceedings of the IEEE/CVF Conference on Computer Vision and Pattern Recognition}, pages 18300--18309, 2023.

\bibitem[Wu et~al.(2021)Wu, Hu, Sheng, and Xu]{wu2021styleformer}
Xiaolei Wu, Zhihao Hu, Lu Sheng, and Dong Xu.
\newblock Styleformer: Real-time arbitrary style transfer via parametric style composition.
\newblock In \emph{Proceedings of the IEEE/CVF International Conference on Computer Vision}, pages 14618--14627, 2021.

\bibitem[Wu et~al.(2020)Wu, Liu, Lin, Lin, and Han]{2020Lite}
Zhanghao Wu, Zhijian Liu, Ji Lin, Yujun Lin, and Song Han.
\newblock Lite transformer with long-short range attention.
\newblock \emph{arXiv}, 2020.

\bibitem[Xie et~al.(2021)Xie, Wang, Yu, Anandkumar, Alvarez, and Luo]{xie2021segformer}
Enze Xie, Wenhai Wang, Zhiding Yu, Anima Anandkumar, Jose~M Alvarez, and Ping Luo.
\newblock Segformer: Simple and efficient design for semantic segmentation with transformers.
\newblock \emph{Advances in Neural Information Processing Systems}, 34:\penalty0 12077--12090, 2021.

\bibitem[Xiong et~al.(2020)Xiong, Yang, He, Zheng, Zheng, Xing, Zhang, Lan, Wang, and Liu]{xiong2020layer}
Ruibin Xiong, Yunchang Yang, Di He, Kai Zheng, Shuxin Zheng, Chen Xing, Huishuai Zhang, Yanyan Lan, Liwei Wang, and Tieyan Liu.
\newblock On layer normalization in the transformer architecture.
\newblock In \emph{International Conference on Machine Learning}, pages 10524--10533. PMLR, 2020.

\bibitem[Zheng et~al.(2021)Zheng, Lu, Zhao, Zhu, Luo, Wang, Fu, Feng, Xiang, Torr, et~al.]{zheng2021rethinking}
Sixiao Zheng, Jiachen Lu, Hengshuang Zhao, Xiatian Zhu, Zekun Luo, Yabiao Wang, Yanwei Fu, Jianfeng Feng, Tao Xiang, Philip~HS Torr, et~al.
\newblock Rethinking semantic segmentation from a sequence-to-sequence perspective with transformers.
\newblock In \emph{Proceedings of the IEEE/CVF conference on computer vision and pattern recognition}, pages 6881--6890, 2021.

\bibitem[Zhou et~al.(2022)Zhou, Huang, Fang, Fu, and Liu]{zhou2022pan}
Man Zhou, Jie Huang, Yanchi Fang, Xueyang Fu, and Aiping Liu.
\newblock Pan-sharpening with customized transformer and invertible neural network.
\newblock In \emph{Proceedings of the AAAI Conference on Artificial Intelligence}, pages 3553--3561, 2022.

\end{thebibliography}
}
\clearpage
\setcounter{page}{1}

\maketitlesupplementary

\section{Feature Extractor}
Figure 1 is an introduction to the architecture of two feature extractors.

\begin{figure}[ht]
    \centering
    \includegraphics[width=1.0\linewidth]{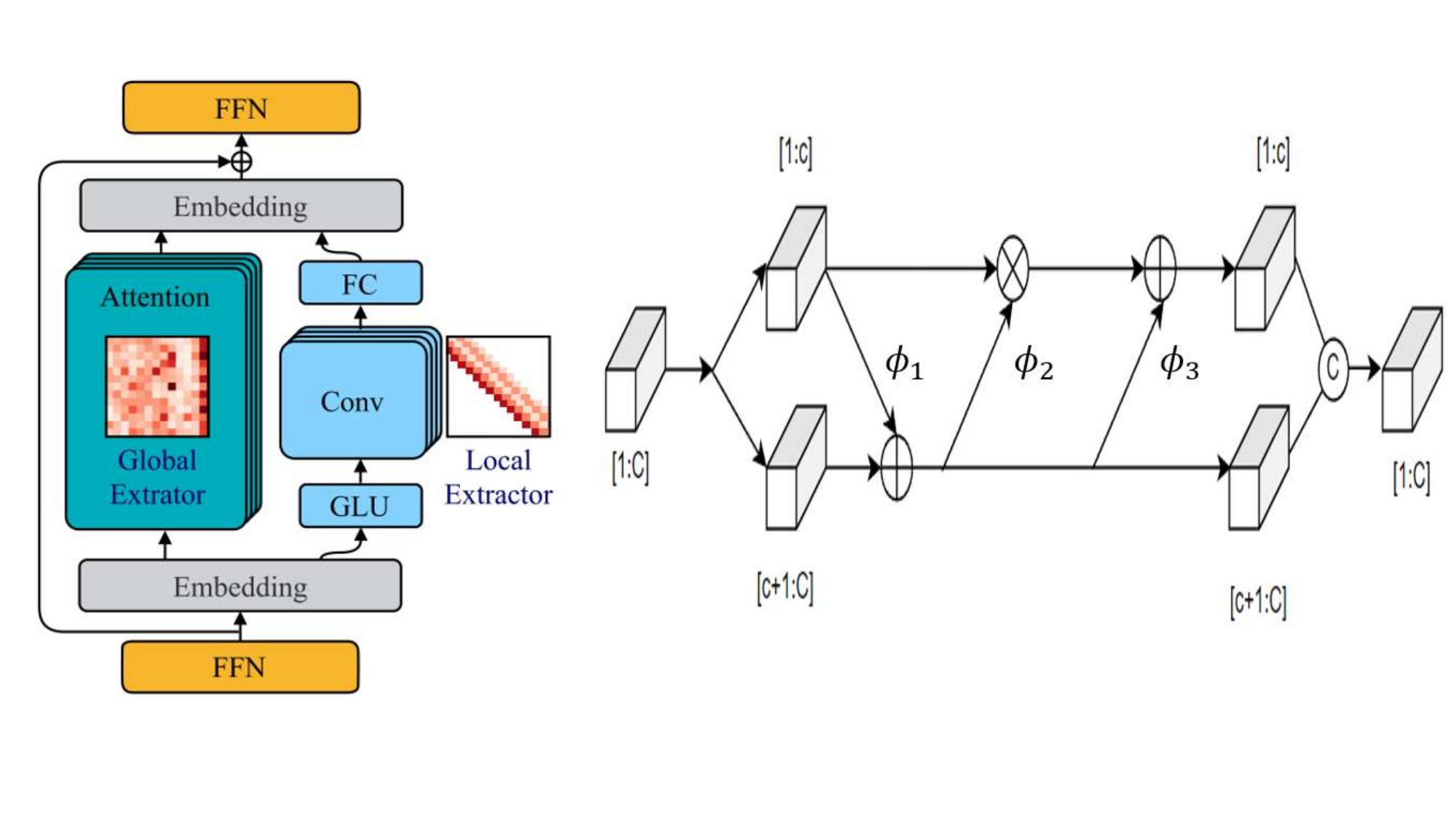}
    \caption{On the left is the workflow of the style extractor, and on the right is the workflow of the content extractor. }
    \label{fig:extractor}
    \vspace{-1em}
\end{figure}

Here, [1:c] represents the $1$st to the $c$th channels. $\phi_i(i=1,2,3)$ are the arbitrary mapping functions. To balance the feature extraction ability and computational consumption, we employ bottleneck residual block (BRB) block in MobileNetV2 \cite{sandler2018mobilenetv2}.

\section{Different Initialization Results}

\begin{figure}[ht]
    \centering
    \includegraphics[width=1.0\linewidth]{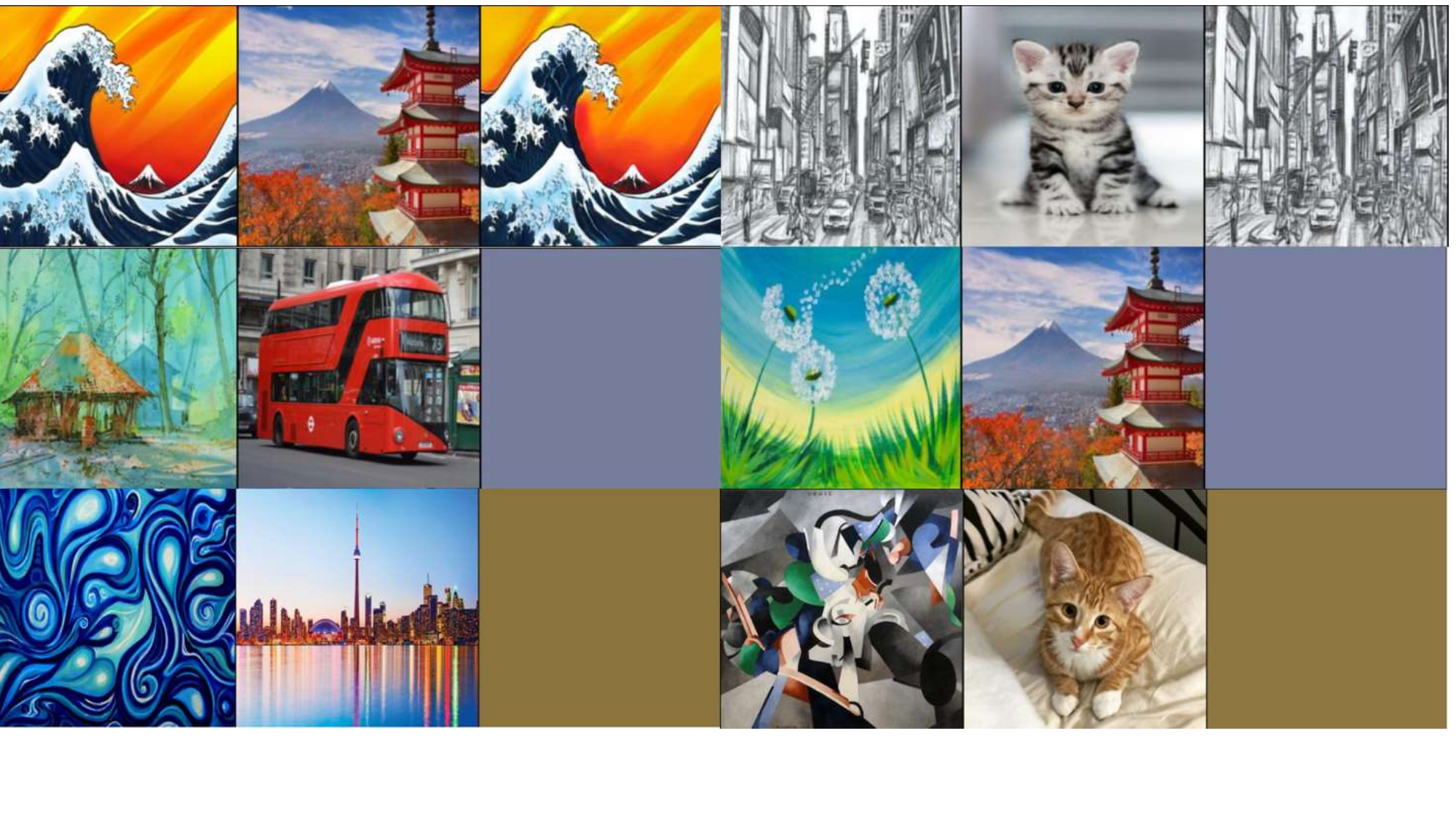}
    \caption{The first row shows the result of using style image for initialization, the second row shows the result of using zero initialization, and the third row shows the result of using random initialization. In each triplet, it is in the order of the style image,  the content image, and the stylization result.}
    \label{fig:supply}
    \vspace{-1em}
\end{figure}

Figure 2 shows the result images generated by initializing $\varepsilon_o$ with style images, zero values, and random values.
It is easy to see that when using different methods to initialize $\varepsilon_o$, our model will produce different qualitative results. Using style images for initialization will make result images closer to style images, and using zero or random values for initialization will make it difficult for us to obtain visually plausible results.
Therefore, we believe $\varepsilon_o$ is the basis for style transfer in our proposed model, and the calculation results of the attention mechanism determine the stylization way of each patch.

\section{Limitation}

Though visually better transfer results have been yeilded, our model still has the drawback of content leak \cite{an2021artflow} like most existing algorithms. After multiple rounds of style transfer for a set of images, some details of the content image will still be lost. We believe that stylization will disrupt the content features we obtain through the content extractor, such as lines, resulting in fewer and fewer extracted content features. We demonstrate this phenomenon in Figure 3.
\begin{figure}[ht]
    \centering
    \includegraphics[width=1.0\linewidth]{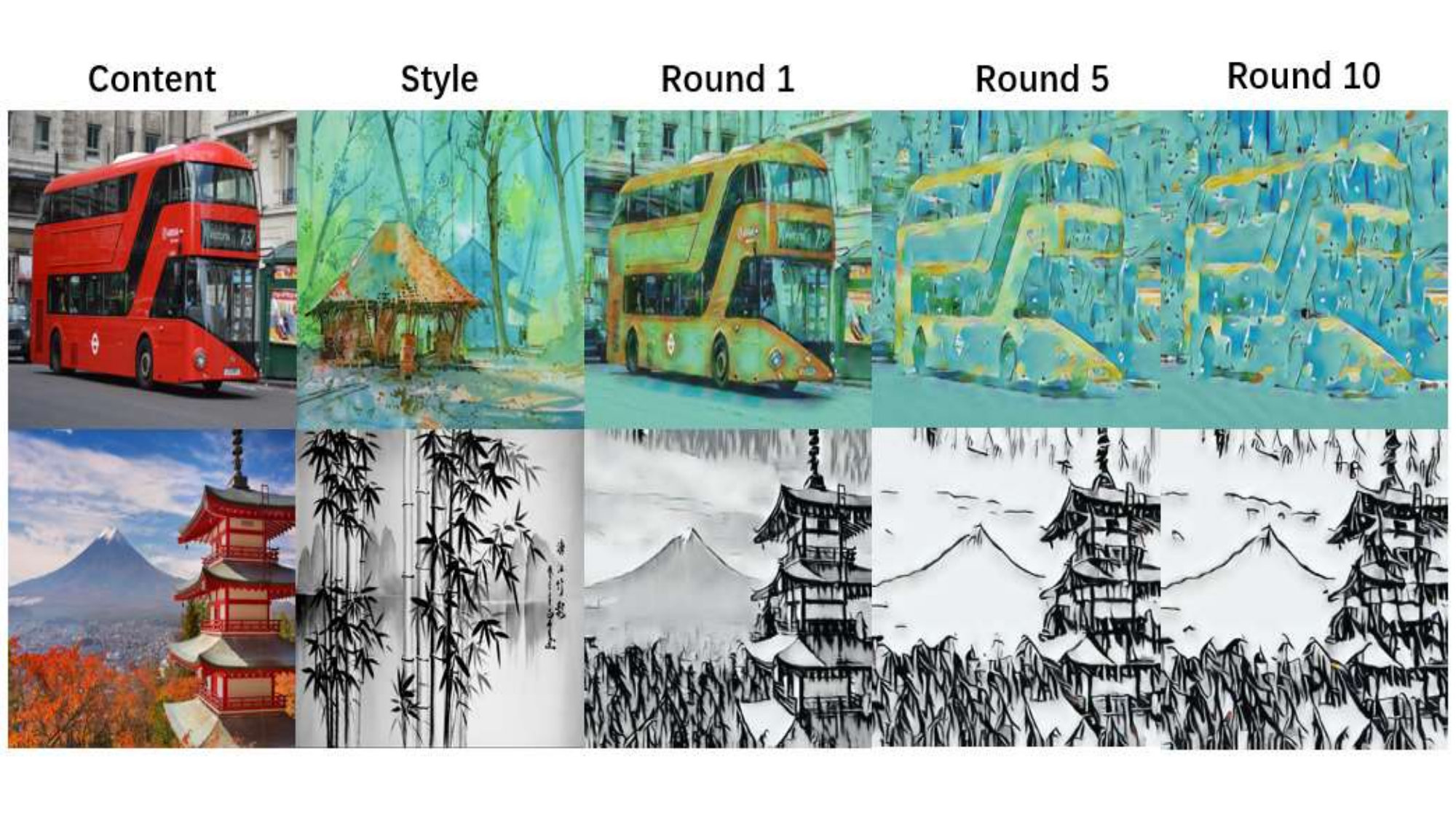}
    \caption{We present the results of stylization for 1, 5, and 10 rounds, respectively.}
    \label{fig:leak}
\end{figure}


\end{document}